\documentclass[letterpaper, 10 pt, conference]{ieeeconf}  

\usepackage{booktabs}
\usepackage{color, colortbl}

\usepackage{adjustbox}

\usepackage{perl_acronyms}
\usepackage{graphicx}
\usepackage{amsmath}
\usepackage{adjustbox}
\usepackage{float}
\usepackage{listings}
\usepackage{subcaption}
\usepackage{hhline}
\usepackage{booktabs}
\usepackage[table]{xcolor}
\usepackage{booktabs}
\usepackage{algorithm}
\usepackage{algpseudocode}
\usepackage{hyperref} 
\usepackage{cleveref}
\definecolor{Gray}{gray}{0.9}
\usepackage{amssymb} 
\usepackage{wrapfig}

\IEEEoverridecommandlockouts                              

\title{\LARGE \bf
PhotoReg: Photometrically Registering 3D Gaussian Splatting Models
}

\author{Ziwen Yuan$^{1}$, Tianyi Zhang$^{1}$, Matthew Johnson-Roberson$^{1}$ and Weiming Zhi$^{1}$
\thanks{$^{1}$Robotics Institute, School of Computer Science,
        Carnegie Mellon University,
        Pittsburgh, PA, USA.
        {\tt\small Email: wzhi@andrew.cmu.edu}
        }%
}

\begin{document}

\maketitle
\thispagestyle{empty}
\pagestyle{empty}

\begin{abstract}
Building accurate representations of the environment is critical for intelligent robots to make decisions during deployment. Advances in photorealistic environment models have enabled robots to develop hyper-realistic reconstructions, which can be used to generate images that are intuitive for human inspection. In particular, the recently introduced \ac{3DGS}, which describes the scene with up to millions of primitive ellipsoids, can be rendered in real time. \ac{3DGS} has rapidly gained prominence. However, a critical unsolved problem persists: how can we fuse multiple \ac{3DGS} into a single coherent model? Solving this problem will enable robot teams to jointly build \ac{3DGS} models of their surroundings. A key insight of this work is to leverage the {duality} between photorealistic reconstructions, which render realistic 2D images from 3D structure, and \emph{3D foundation models}, which predict 3D structure from image pairs. To this end, we develop PhotoReg, a framework to register multiple photorealistic \ac{3DGS} models with 3D foundation models. As \ac{3DGS} models are generally built from monocular camera images, they have \emph{arbitrary scale}. To resolve this, PhotoReg actively enforces scale consistency among the different \ac{3DGS} models by considering depth estimates within these models. Then, the alignment is iteratively refined with fine-grained photometric losses to produce high-quality fused \ac{3DGS} models. We rigorously evaluate PhotoReg on both standard benchmark datasets and our custom-collected datasets, including with two quadruped robots. The code is released at \url{https://ziweny11.github.io/photoreg}.



\end{abstract}

\section{INTRODUCTION}
\label{sec:introduction}
Constructing representations is a central requirement to enable autonomous robot operations. Robots often carry extroceptive sensors, such as cameras, to observe the environment, and representations are needed to condense this collected information ~\cite{agha2021nebula,keetha2023anyloc, zhang2023nerf, zhang2024darkgs}. In this paper, we focus our attention on \emph{photorealistic reconstruction} models. These models uniquely convert robot sensor inputs into models with which photorealistic images can be rendered. These photorealistic images can better enable non-roboticists to visualize the robot's environment. In particular, recent developments in the area have led to the development of \ac{3DGS}, which has, for the first time, enabled the real-time rendering of photorealistic images from the reconstruction.  This breakthrough has significant implications for robotics, where the ability to visualize complex environments in real-time enhances robot navigation, manipulation, and interaction. 
\begin{figure}[t]%
\centering
\includegraphics[width=0.99\linewidth]{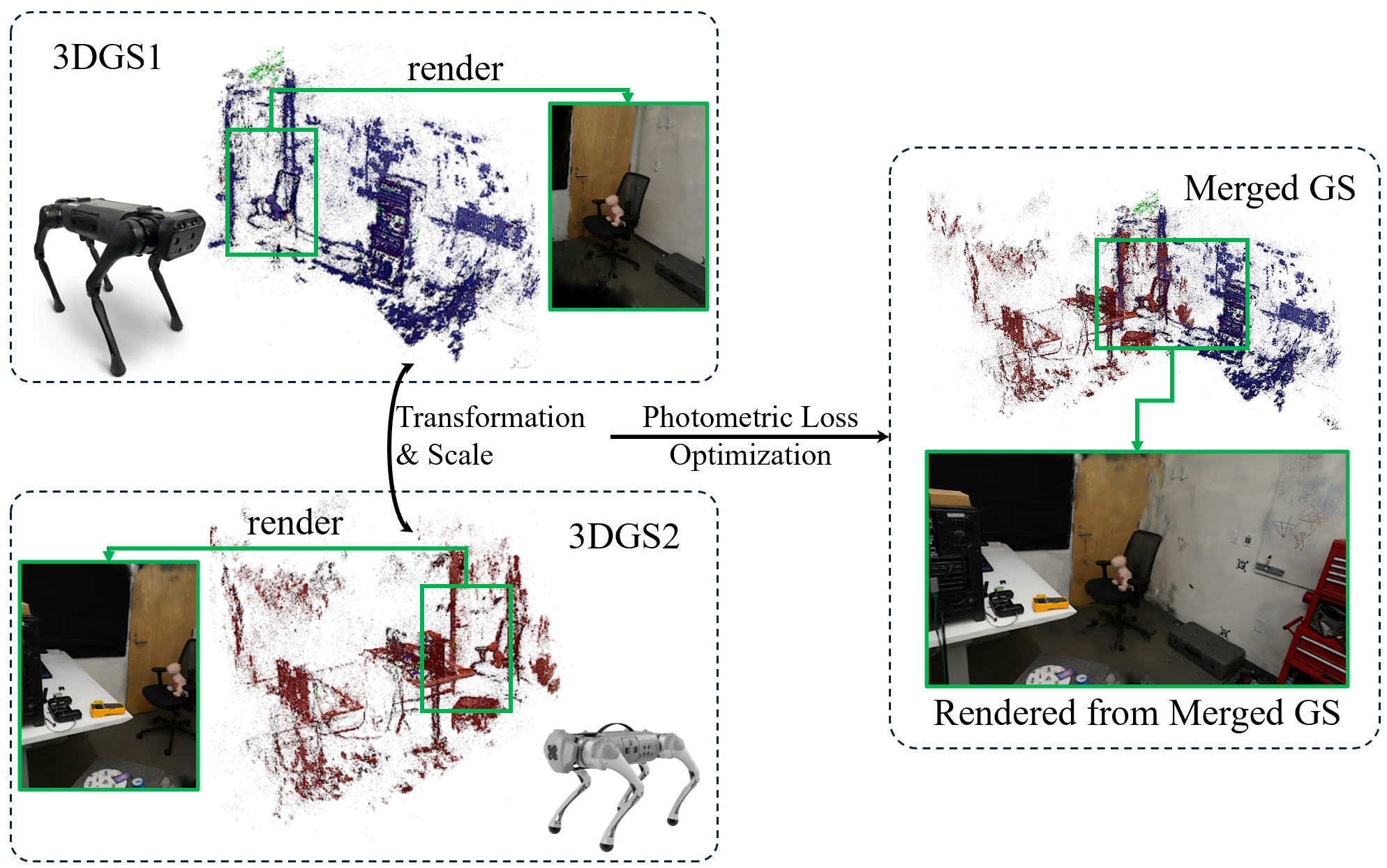}
    \caption{PhotoReg takes two input 3D Gaussian Splatting models and aligns them into a merged Gaussian Splatting model. Transformation and Scale information between two inputs is obtained through 3D visual foundation models and further refined photometrically.}
    \label{method:banner}
\end{figure}

This paper studies the problem of combining multiple \ac{3DGS} models, built separately, into a single unified model. Solving this problem will allow a team of robots to explore and map large unknown spaces in a decentralized manner. A unified model, created by merging the individual representations of each robot, can be distributed to all robots. Relative to sharing all the image data with each robot in the team, distributing the condensed representation only enables efficient usage of bandwidth and latency. To tackle the challenge of \ac{3DGS} fusion, we present our \textbf{Photo}metrical 3D Gaussian \textbf{Reg}istration framework (PhotoReg). Classical registration methods, such as \ac{ICP} and its variants ~\cite{zhang2021self, zhou2016fast, yuan2020self}, focus primarily on aligning point clouds by minimizing the distance between corresponding points  ~\cite{du2007icp}. However, the continuous and complex geometric representations inherent in 3DGS differ significantly from the discrete point sets managed by traditional registration methods, necessitating innovative approaches like PhotoReg for effective alignment.

PhotoReg utilizes \emph{3D foundation models}, trained on Internet-scale datasets, to derive initial 3D structures from 2D image pairs. These models provide rough estimations that facilitate the initial alignment of \ac{3DGS} models, especially in scenarios where the overlap between the models is minimal. As the individual \ac{3DGS} models may not be of the same scale, PhotoReg actively aligns their scales by considering confidence-aware depth estimates in each model. Subsequently, PhotoReg optimizes fine-grained \emph{photometric losses}, which measure the quality of rendered images from the model, to ensure tight alignment between the \ac{3DGS}. We provide extensive empirical evaluations of PhotoReg, both on classic benchmark datasets and on custom-collected data. This includes a custom dataset collected by two quadrupeds operating in a common area.

Concretely, PhotoReg makes the following methodological innovations:
\begin{enumerate}
\item Leverage 3D foundation models to tackle \ac{3DGS} alignment when there is minimal overlap;
\item Resolve scaling disparities in \ac{3DGS} models by computing confidence-aware depth estimates to rescale individual models;
\item Precise tuning of the fused model via optimizing the quality of rendered images.
\end{enumerate}

\section{RELATED WORK}
\label{sec:relatedwork}
\textbf{Photorealistic Reconstructions:} Robots operating in unknown environments require internal representations to understand their surroundings, to effectively generate motions \cite{PDMP,GeoFab_gloabL_opt,Diff_templates}. Traditionally, this has been maps of occupancy \cite{OccupancyGridMaps, HM} or dynamics \cite{DirectionalGridMaps, sptemp, OTNet, KTM}. Developments in neural networks have led to \ac{NeRF} ~\cite{mildenhall2020nerf} which learn photorealistic 3D scenes, where images from novel views can be rendered. Subsequent extensions sped up training \cite{mueller2022instant}. However, rendering was often slow. Recent work on \ac{3DGS} ~\cite{kerbl3Dgaussians} proposes to model the scene as a mixture of 3D Gaussians which enables real-time photorealistic rendering. Extensions to \ac{3DGS} \cite{zhang2024darkgs, zhang2024recgs} have improved the robustness of the method.

\textbf{3D Registration:} In robot perception, registration refers to finding the transformation between two 3D structures. Registering two point clouds has been widely studied over time. \ac{ICP}~\cite{chen1992objectICP} alternatively finds the pairs of correspondence points and estimate the rigid body transformation between them, based on the closest-point assumption. Variants such as color ICP ~\cite{park2017colored}, Point to Plane ICP ~\cite{besl1992method}, and Robust ICP ~\cite{jian2011robust} have improved the method in terms of accuracy and efficiency.  
Methods that register two NeRFs have been explored.
NeRF2NeRF~\cite{goli2023nerf2nerf} proposes to align two \ac{NeRF}s by manually selecting key points.
DReg-NeRF~\cite{DReg2023} further advances 3D registration by automating the alignment of NeRF models using deep learning. An attempt has been made to explore \ac{3DGS} registration: LoopSplat~\cite{zhu2024loopsplat} introduces a novel loop closure technique by registering 3D Gaussian splats. However, LoopSplat relies on RGB-D images for depth sensor readings, which limits its applicability when depth sensors are unavailable or unreliable. PhotoReg enables the registration of \ac{3DGS} in the absence of depth sensors.

\textbf{Visual Foundation Models:} Robotics benefits from transformer models trained on internet-scale data \cite{Bommasani2021FoundationModels}. Particularly for robot perception, such visual foundation models include ~\cite{chen2020simple, donahue2019large, grill2020bootstrap}. These models act as plug-and-play modules to facilitate a range of downstream tasks. For example, DINOv2~\cite{oquab2023dinov2} was trained on internet-scale unlabeled data, employing self-supervision techniques that allow it to develop a deep understanding of visual content without the need for explicit annotations. DUSt3R~\cite{DUSt3R_cvpr24} is a 3D foundation model used in PhotoReg. It is designed to generate 3D pointmaps from RGB images, enabling pose estimation, and has been applied to downstream robot manipulator perception \cite{JCR, zhi20243d}. PhotoReg leverages the emergent capabilities of foundation models to perform robust alignment.

\section{PRELIMINARIES: Foundation Models}
\label{sec:preliminaries}
This work makes use of foundation models, which are large deep-learning models trained on internet-scale datasets. These models are intended as plug-and-play modules, used to facilitate a range of downstream tasks without retraining on specific datasets. In this section, we briefly outline two foundation models used in our PhotoReg framework: DUSt3R and DINOv2. More details are available in the original papers, \cite{DUSt3R_cvpr24} and \cite{oquab2023dinov2}.

\begin{figure}[t]%
\centering
\includegraphics[width=0.95\linewidth]{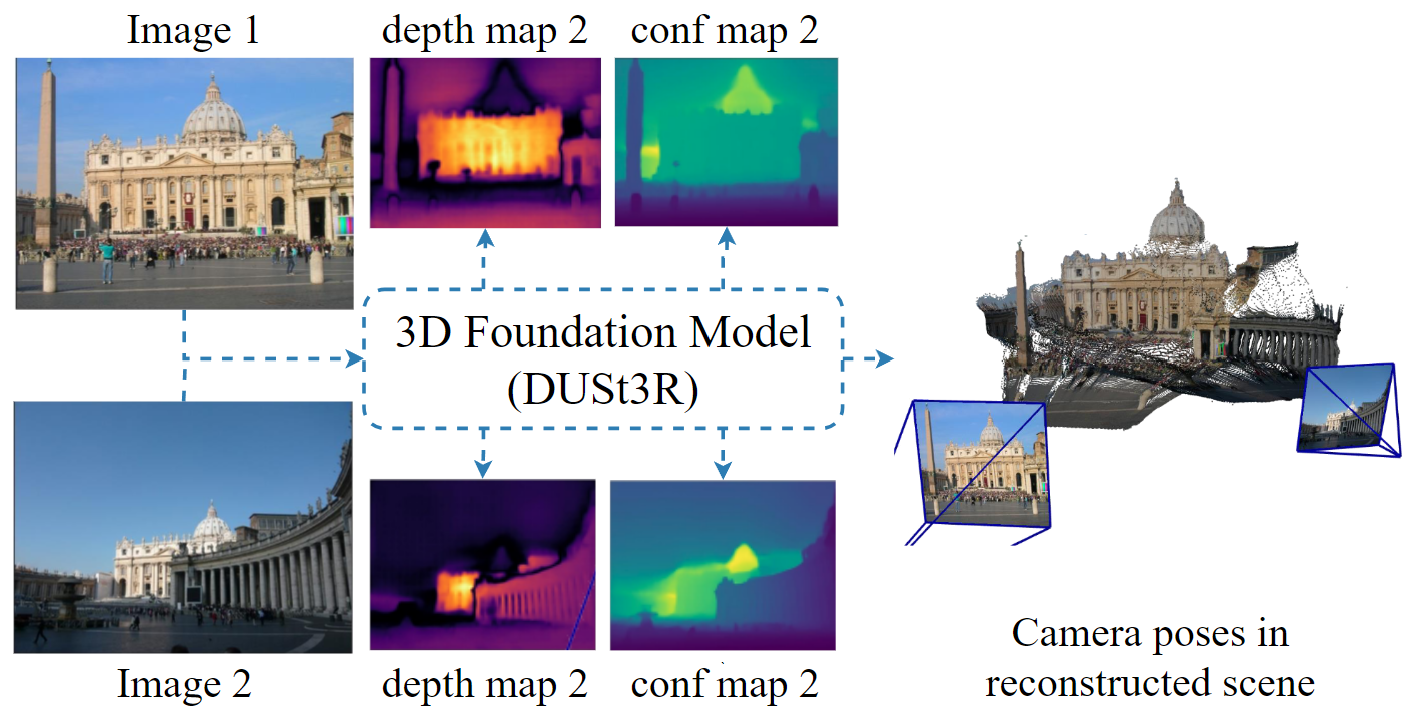}
    \caption{DUSt3R workflow: Input are two RGB images, output are corresponding depth maps, confidence maps, and a 3D reconstructed scene with camera poses of input images recovered.}
    \label{prelim:DUSt3R}
\end{figure}

\begin{figure}[t]%
\centering
\includegraphics[width=0.95\linewidth]{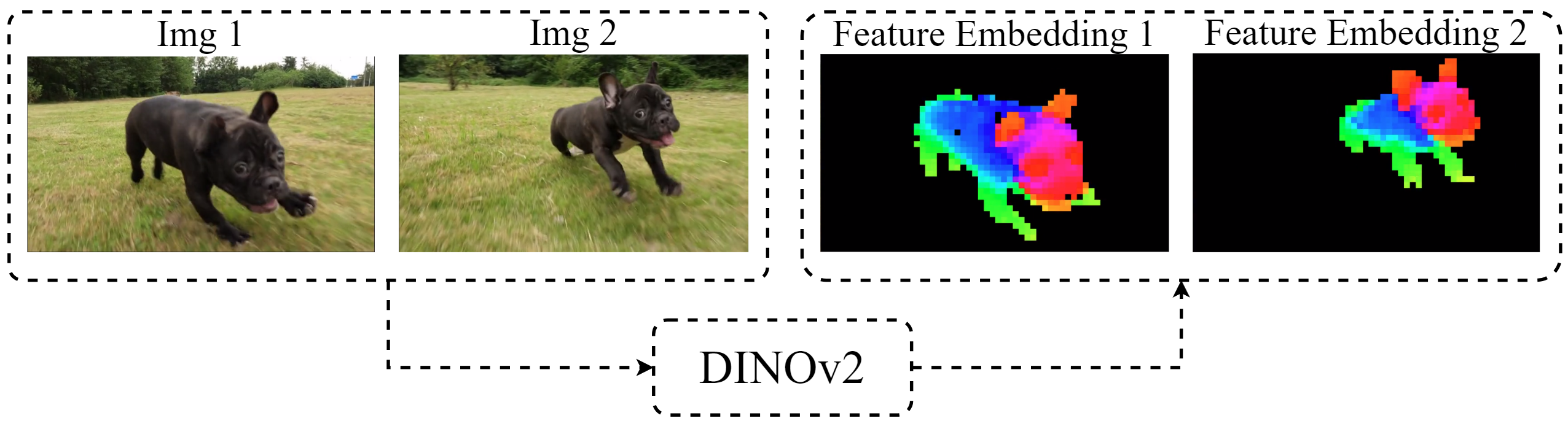}
    \caption{DINOv2 Workflow: Input images are transformed into feature embeddings, with the first three principal components visualized in RGB values. This figure demonstrates that similar objects are mapped to closely related embeddings, and are invariant to differences in position or angle.}
    \label{prelim:dinov2}
\end{figure}
\textbf{DUSt3R:} At the core of DUSt3R is a large vision transformer~\cite{dosovitskiy2020vit}. It takes as input 2 RGB images of width $W$ and height $H$, $I_1, I_2 \in {\mathbb{R}}^{W \times H \times 3}$ and outputs 2 corresponding 3D pointmaps \(X_{1,1}, X_{2,1} \in {\mathbb{R}}^{W \times H \times 3}\) with associated confidence maps \(C_{1}, C_{2} \in {\mathbb{R}}^{W \times H}\) and depth maps $D_{1}, D_{2} \in \mathbb{R}^{W \times H}$, from which it further recovers a variety of geometric quantities, such as relative camera poses and fully-consistent 3D reconstruction. In our proposed PhotoReg framework, we will make use of the above outputs. This workflow is illustrated in~\cref{prelim:DUSt3R}. As DUSt3R is entirely data-driven, it does not need to identify hand-crafted features within our images to find correspondence. As a result, it is capable of accurately finding relative camera poses, even when visual overlap between the inputted image pair is minimal. Our PhotoReg framework capitalizes on this feature, enabling the alignment of GS models with minimal overlap.


\textbf{DINOv2}~\cite{oquab2023dinov2}: DINOv2 is a \emph{visual foundation model} that employs a transformer model, trained in a self-supervised manner over extensive image datasets. It takes as input a single image and outputs a corresponding vector embedding. These embeddings are generally invariant to spatial transformation, with semantically similar objects being close in this embedding space. We use DINOv2 to search for potentially adjacent regions within each \ac{3DGS} model. These adjacent regions exhibit overlapping visual features such as common elements identifiable in both images. DINOv2 helps to detect these overlapping features and guides the selection of input images into the DUSt3R model discussed above. An overview of the workflow of DINOv2 is shown in \cref{prelim:dinov2}.

\section{METHODOLOGY}
\label{sec:methodology}
\begin{figure}[t]%
\centering
\includegraphics[width=0.99\linewidth]{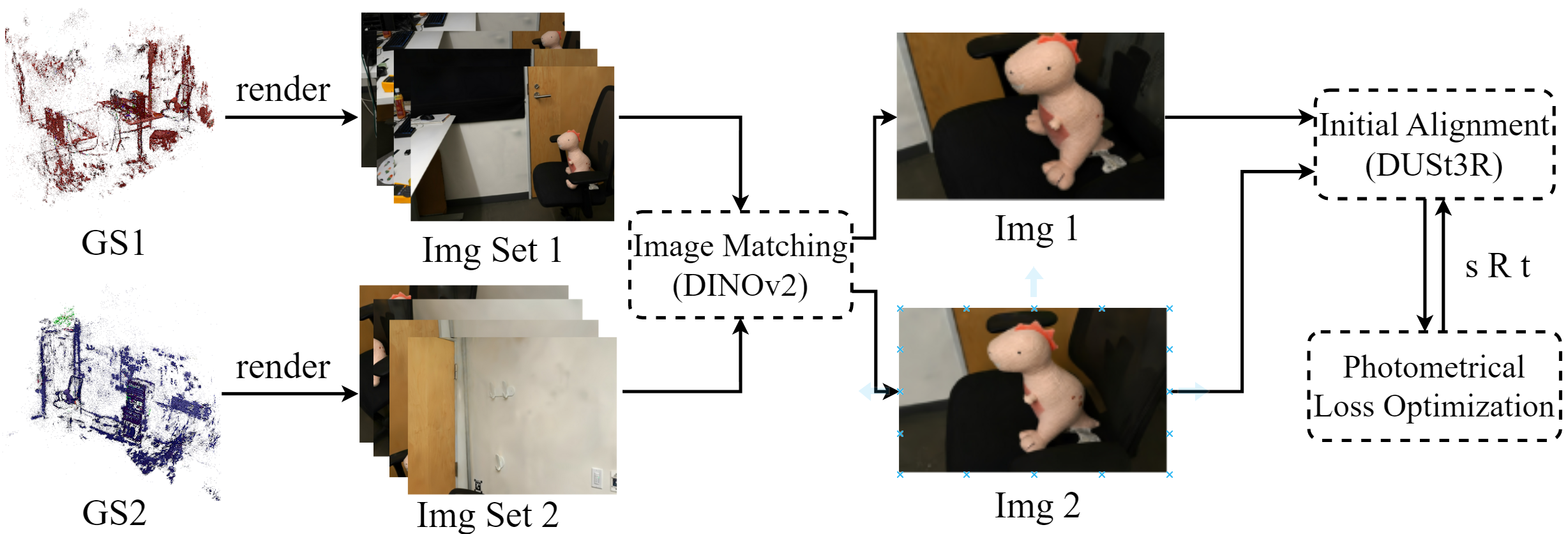}
    \caption{This workflow starts by rendering images from training poses of two overlapping 3D Gaussians. We then select images where the overlap is detected using a visual foundation model. The initial pose between these images is established based on rendered depth maps and a 3D foundation model. By optimizing the photometric loss, we refine the rigid body transformation and scale ratio between the two 3DGS, achieving an aligned Gaussian model.}
    \label{method:wf}
\end{figure}

\subsection{Problem Setup}
The primary challenge addressed in this work involves the fusion of \ac{3DGS} models. Specifically, given input \ac{3DGS} models, \(G_1\) and \(G_2\), our proposed method aims to discover a transformation function \(T\) that cohesively aligns \(G_2\) to \(G_1\) within the coordinate frame of \(G_1\). As \ac{3DGS} are of arbitrary scale, $T$ needs to handle $G_{1}$ and $G_{2}$ that may be of vastly differing scales. Before delving into the details of our proposed method, PhotoReg, we first provide a mathematical definition of \ac{3DGS} models and the corresponding transformation functions.

\subsection{Notation and Transforming Gaussian Splats}
A \ac{3DGS} model $G$ consists of a set of 3D Gaussians, where each Gaussian is defined by its 3D position, \(\boldsymbol{\mu}\); a covariance matrix, \(\Sigma\), describing the spread and orientation of the Gaussian distribution in 3D space; opacity, \(\alpha\); and Spherical Harmonics (SH) coefficients $c$, containing color information:
\begin{equation}
G = \{(\boldsymbol{\mu}, \Sigma, \alpha, c)\}. 
\end{equation}
\textbf{Transformation}:
The transformation of a \ac{3DGS} model involves applying scaling, rotation, and translation to each attribute of each Gaussian. Let \(T_A^B\) denote the transformation function that maps a \ac{3DGS} model from some general coordinate frame \(A\) to coordinate frame \(B\). This function takes as input \([G]_A\), the \ac{3DGS} model in coordinate frame \(A\), and outputs \([G]_B\), the corresponding \ac{3DGS} model in frame \(B\) after the transformation, denoted as$[G]_B = T_A^B([G]_A)$.

The transformation function \(T_A^B\) can be composed by scaling factor $s\in \mathbb{R}$, rotation $R \in SO(3)$, and translation $t\in \mathbb{R}^3$, respectively. Suppose \([G]_A = \{(\boldsymbol{\mu}_A, \Sigma_A, \alpha_A, c_A)\}\) and \([G]_B = \{(\boldsymbol{\mu}_B, \Sigma_B, \alpha_B, c_B)\}\). The transformation mapping for each attribute is defined as follows:
\begin{itemize}
    \item \textbf{3D Position}: \(\boldsymbol{\mu}_B = sR  \boldsymbol{\mu}_A + t\).
    \item \textbf{Covariance Matrix}: \(\Sigma_B = R  \Sigma_A\).
    \item \textbf{Opacity}: \(\alpha_B = \alpha_A\) (Spatially invariant).
    \item \textbf{Spherical Harmonics (SH) Coefficients}: The SH coefficients are translation-invariant and transformed by:
    \[
    c_B = D(R, \text{order}) \cdot c_A.
    \]
    where \(D(R, \text{order})\) is the Wigner D-matrix \cite{gaussiansplatl2024} corresponding to the rotation for a known specific order.
\end{itemize}


\subsection{PhotoReg Overview}
PhotoReg consists of four sequential stages: 
\textbf{Foundational Image Matching}: Select rendered image pairs of adjacent regions in each input \ac{3DGS} model as input into the 3D foundation model; \textbf{Initial Estimation}: Obtain an initial estimation of rotation and translation for alignment through the 3D foundation model; \textbf{Scale Estimation}: Resolve scale discrepancies between the \ac{3DGS} models through confidence-weighted depth maps; \textbf{Optimization}: Optimize scale, rotation, and translation simultaneously through photometric loss minimization.
The workflow is illustrated in \cref{method:wf} and further discussed in the following subsections.

\subsection{Foundational Image Matching}
Here, we detail the Foundational Image Matching (FIM) procedure, which utilizes visual foundation models, notably DINOv2, to initiate our registration process. The input to the FIM process consists of two \ac{3DGS} Models, denoted \(G_1\) and \(G_2\), each associated with a set of camera poses, \(C_1\) and \(C_2\). The primary aim of FIM is to identify and extract two high-quality rendered images at given camera poses from \(G_1\) and \(G_2\) which are similar at a semantic level, and invariant to viewing poses. That is, two images of the same object but at vastly differing angles will be identified as similar.
To select appropriate images as input into the 3D foundation model, we first render image sets \(I_1\) and \(I_2\) from \(G_1\) and \(G_2\) at diverse poses. We aim to select a suitable image pair \(img_1 \in I_1\) and \(img_2 \in I_2\) for coarse registration. 
We seek $v_{1} \in V_1$ and $v_{2} \in V_2$
such that
\begin{equation}
\operatorname{argmax}_{v_{1} \in V_1, v_{2} \in V_2} \cos(v1, v2)
\end{equation}
This approach identifies pairs of images, with one generated from $G_{1}$ and the other from $G_{2}$, which are semantically and visually similar.

\subsection{Coarse Registration via 3D Foundation Models}
We proceed to use the image pair $(img_1, img_2)$ as input into a 3D foundation model, DUSt3R \cite{DUSt3R_cvpr24}, to obtain an initial coarse registration, to approximately align \(G_2\) into the coordinate frame of \(G_1\). We input \(img_1\) and \(img_2\) into DUSt3R, which estimates a rigid transformation \(T_{img_{2}}^{img_{1}}\), with rotation and translation. However, the scaling factor between \(G_1\) and \(G_2\) remains unknown: DUSt3R takes 2D images as inputs, making it impossible to recover the scaling directly.

\begin{figure}[t]
\centering
\includegraphics[width=0.7\linewidth]{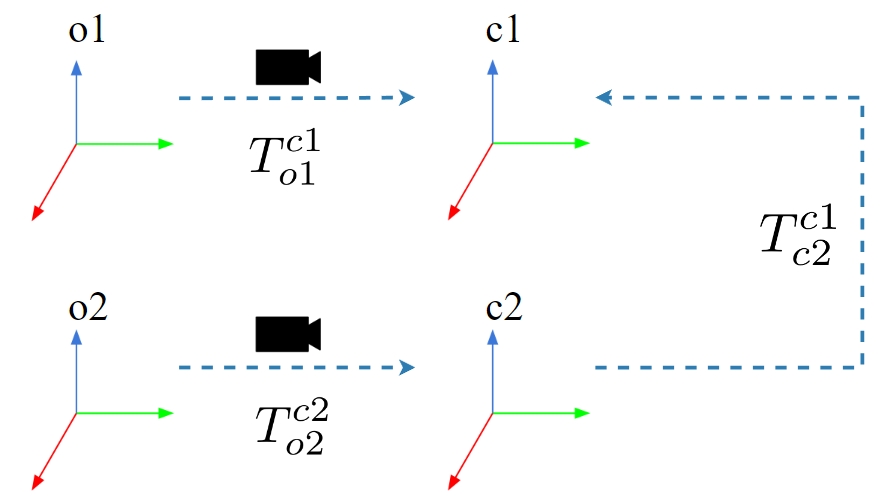}
\caption{Sequence of transformations: \(o_1\) and \(o_2\) are the coordinate frames of \(G_1\) and \(G_2\); \(c_1\) and \(c_2\) are the camera coordinate frames of \(G_1\) and \(G_2\). $T^{c1}_{c2}$ is not available as the scale is unknown.}
\label{method:coordinate}
\end{figure}

The next step involves applying the transformations obtained from the 3D foundation model back to the original 3DGS models. This process involves a sequence of transformation steps, as illustrated in \cref{method:coordinate}. We define the coordinate frames as follows: For a coordinate frame \(P\), \([G_i]_P\) denotes the 3DGS model \(G_i\) under the coordinate frame \(P\). Specifically, \(P_{oi}\) is the original frame of the 3DGS model \(G_i\), and \(P_{ci}\) is the coordinate frame of the camera to produce image $img_{i}$. The original frames of the splatting models are constructed arbitrarily.

Our objective is to obtain a transformation function \(T_{o2}^{o1}\) that directly transforms \(G_2\) into the coordinate frame of \(G_1\), resulting in:
\begin{equation}
[G_2]_{P_{o1}} = T_{o2}^{o1}([G_2]_{P_{o2}}),
\end{equation}

We follow the sequence of transformations in \cref{method:coordinate}. Here, \(T^{c1}_{o1}\) and \(T^{c2}_{o2}\) are the world-to-camera transformations, corresponding to the images $img_{1}$ and $img_{2}$, which are known. The transformation between the two camera poses is denoted as $T^{c1}_{c2}$, which consists of rotation, translation and scaling \((R_{c2}^{c1}, t_{c2}^{c1}, s_{c2}^{c1})\). We highlight that $R_{c2}^{c1}, t_{c2}^{c1}$ match the output from DUSt3R, \(T_{img_{2}}^{img_{1}}\). However, the scale $s_{c2}^{c1}$ is unknown. Following the composition of transformations, we have 
\begin{equation}
[G_2]_{P_{o1}} = \left((T^{c1}_{o1})^{-1} \circ T^{c1}_{c2} \circ T^{c2}_{o2}\right)([G_2]_{P_{o2}}) \label{transformeq}
\end{equation}

 The only unknown parameter is the scale ratio \(s_{c2}^{c1}\). As both the world-to-camera transformations have known absolute scales, $s_{c2}^{c1}$ is the difference in scale between the two \ac{3DGS} models. That is, \(s_{c2}^{c1} = s_{o2}^{o1}\). In the following subsection, we will estimate this final unknown parameter, \(s_{o2}^{o1}\), the scale ratio between, $G_1$ and $G_2$.

\subsection{Scale Estimation}

In this step, we estimate the scale ratio \(s_{o2}^{o1}\) by comparing depth maps at corresponding camera poses across different coordinate frames. A depth map can be represented as a two-dimensional matrix, where each element indicates the distance from the viewpoint (e.g., a camera) to a point in the scene along the line of sight. 

From a given camera pose, depth maps can be extracted from standard \ac{3DGS} models \cite{zhang2024radegs}, and are also outputted by DUSt3R, during the initial coarse alignment. Pixel-wise confidence maps for the depth maps are also outputted. Here, we denote the depth maps for $img_{1}$ and $img_{2}$ obtained from their respective \ac{3DGS} models as $D_{1}, D_{2}\in\mathbb{R}^{W \times H}$, and the depth maps from DUSt3R with the same images as $\Bar{D}_{1}, \Bar{D}_{2}\in\mathbb{R}^{W \times H}$, along with pixel-wise confidence maps $C_{1}, C_{2} \in\mathbb{R}^{W \times H}$. Here, $W$ and $H$ denote the image width and height. 

A key insight is that the depths $\Bar{D}_{1}$ and $\Bar{D}_{2}$ are in the same coordinate system and of the same scale. We can then estimate a confidence-weighted scale between $G_{1}$ and $G_{2}$,

\begin{align}
s_{o2}^{o1}=\frac{\sum(C_{1}\otimes(D_{1}/\Bar{D}_{1}))}{\sum(C_{2}\otimes(D_{2}/\Bar{D}_{2}))},
\end{align}
where $\otimes$ denotes the element-wise product. With the estimate of scale $s^{o1}_{o2}$, we have an initial transformation $T^{c1}_{c2}$ which roughly aligns $G_{1}$ and $G_{2}$. We now shift our focus to refining the alignment. 

\subsection{Precise Refinement via Photometric Optimization}
After roughly aligning $G_1$ and $G_2$, we further refine the alignment by rendering images at a novel pose, $\Bar{C}$, from both $G_{1}$ and $G_{2}$. Then, we minimize the photometric loss between the rendered images, and optimize with respect to our transformation parameters. We note that the differentiability of \ac{3DGS} models and rendering enables gradient-based optimization to be propagated back to the transformation parameters. The loss used computes the $L1$ distance, masked by binary indicators whether there is anything rendered at a given pixel, between the rendered images, at the same pose, but from $G_{1}$ and $G_{2}$. Specifically,
\begin{align}
L=l1_{masked}(F(G_{1},\Bar{C}),F(G_{2},\Bar{C}), M_{1}\otimes M_{2}),
\end{align}
where $F$ is the differentiable rendering function that generates an image given camera pose and the \ac{3DGS} model. The masked $L1$ distance, $l1_{masked}$, gives the $L1$ distance masked by an element-wise binary mask. The element-wise binary mask used is $M_{1}\otimes M_{2}$, where $M_{1}$ and $M_{2}$ are binary masks indicating whether anything has been rendered at each pixel.  

We differentiate $L$ with respect to the parameters of $T^{c1}_{c2}$, given as \(s_{o2}^{o1}\), \(R_{c2}^{c1}\), \(t_{c2}^{c1}\). We use gradient-based optimizers to minimize the loss and perform the detailed alignment.

\section{EXPERIMENTS}
\label{sec:experiments}
\begin{table}[t]
\centering
\caption{\small Performance metrics (SSIM, PSNR, LPIPS) across different scenes. We use original \ac{3DGS} as the ground truth. Our method approaches the ground truth quality outperforming ICP, and COLMAP baselines.}
\label{tableresult}

\begin{adjustbox}{width=1\linewidth}
\begin{tabular}{@{}cccc@{}}
\toprule
\multicolumn{4}{c}{\textbf{Collected Datasets}} \\
\midrule
\textbf{Metric} & \textbf{SSIM} & \textbf{PSNR (dB)} & \textbf{LPIPS} \\
\midrule
\multicolumn{4}{c}{\textbf{Workroom 1 (Low Overlap)}} \\
\rowcolor{Gray}
\textbf{GT} & 0.969 & 31.6 & 0.092 \\
\textbf{Ours} & \textbf{0.900} & \textbf{28.6} & \textbf{0.238} \\
\textbf{ICP} & \multicolumn{3}{c}{Diverged} \\
\textbf{COLMAP} & \multicolumn{3}{c}{Diverged} \\
\midrule
\multicolumn{4}{c}{\textbf{Workroom 2 (Moderate Overlap)}} \\
\rowcolor{Gray}
\textbf{GT} & 0.952 & 30.4 & 0.098 \\
\textbf{Ours} & \textbf{0.906} & \textbf{28.4} & \textbf{0.196} \\
\textbf{ICP} & 0.771 & 18.3 & 0.311 \\
\textbf{COLMAP} & 0.870 & 21.1 & 0.204 \\
\midrule
\multicolumn{4}{c}{\textbf{Sofa}} \\
\rowcolor{Gray}
\textbf{GT} & 0.868 & 22.4 & 0.147 \\
\textbf{Ours} & \textbf{0.803} & \textbf{19.9} & \textbf{0.236} \\
\textbf{ICP} & 0.763 & 17.6 & 0.269 \\
\textbf{COLMAP} & \multicolumn{3}{c}{Diverged} \\
\bottomrule
\end{tabular}
\hfill 
\begin{tabular}{@{}cccc@{}}
\toprule
\multicolumn{4}{c}{\textbf{Datasets used in \cite{kerbl3Dgaussians}}} \\
\midrule
\textbf{Metric} & \textbf{SSIM} & \textbf{PSNR (dB)} & \textbf{LPIPS} \\
\midrule
\multicolumn{4}{c}{\textbf{Playroom}} \\
\rowcolor{Gray}
\textbf{GT} & 0.913 & 30.0 & 0.244 \\
\textbf{Ours} & \textbf{0.857} & \textbf{25.6} & \textbf{0.297} \\
\textbf{ICP} & 0.662 & 13.7 & 0.526 \\
\textbf{COLMAP} & 0.402 & 12.6 & 0.543 \\
\midrule
\multicolumn{4}{c}{\textbf{Truck}} \\
\rowcolor{Gray}
\textbf{GT} & 0.882 & 25.1 & 0.161 \\
\textbf{Ours} & \textbf{0.739} & \textbf{19.5} & \textbf{0.290} \\
\textbf{ICP} & 0.561 & 13.6 & 0.480 \\
\textbf{COLMAP} & 0.765 & 18.9 & 0.410 \\
\midrule
\multicolumn{4}{c}{\textbf{Train}} \\
\rowcolor{Gray}
\textbf{GT} & 0.816 & 21.7 & 0.238 \\
\textbf{Ours} & \textbf{0.669} & \textbf{17.4} & \textbf{0.349} \\
\textbf{ICP} & 0.587 & 12.8 & 0.429 \\
\textbf{COLMAP} & 0.634 & 16.2 & 0.384 \\
\bottomrule
\end{tabular}
\end{adjustbox}
\end{table}

\begin{figure}[t]
\centering
\includegraphics[width=0.99\linewidth]{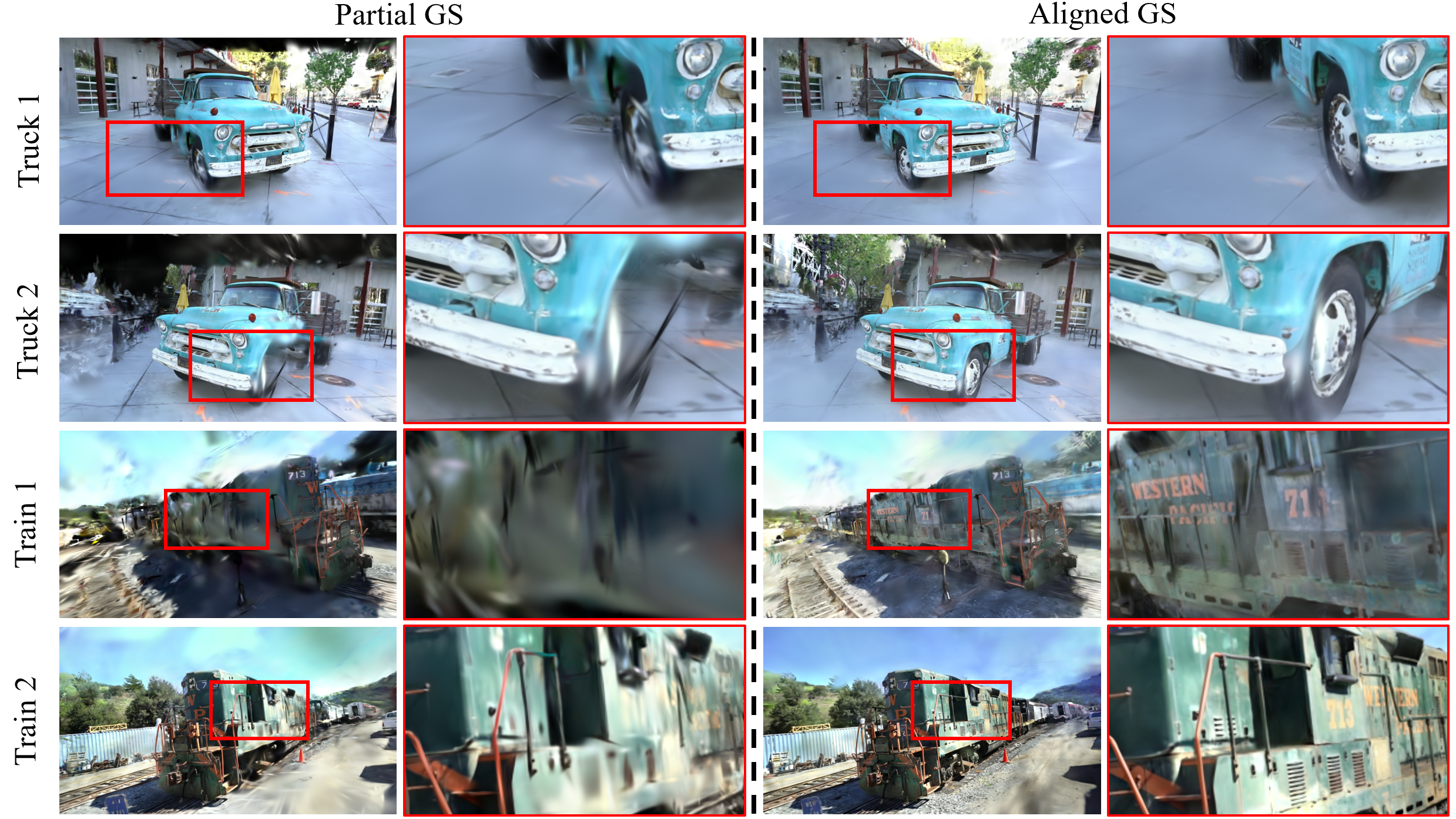}
    \caption{Visualization of PhotoReg registration results. We observe lower image quality in the partial models, particularly in the regions outlined in red. The image qualities of the same regions are drastically better in the aligned models.}
    \label{exp:results1}
\end{figure}

\begin{figure}[t]
    \centering
    
    \begin{subfigure}{0.22\linewidth}
        \centering
        \fbox{\includegraphics[width=\linewidth]{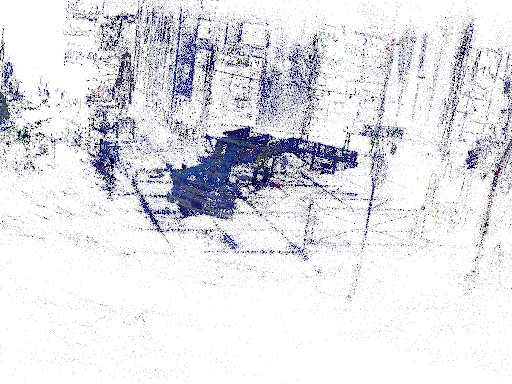}}
    \end{subfigure}
    \hspace{1mm} 
    \begin{subfigure}{0.22\linewidth}
        \centering
        \fbox{\includegraphics[width=\linewidth]{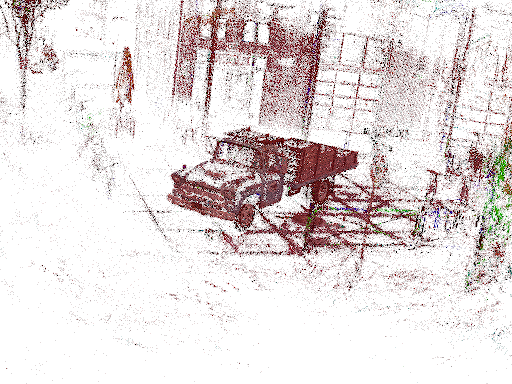}}
    \end{subfigure}
    \hspace{1mm} 
    \begin{subfigure}{0.22\linewidth}
        \centering
        \fbox{\includegraphics[width=\linewidth]{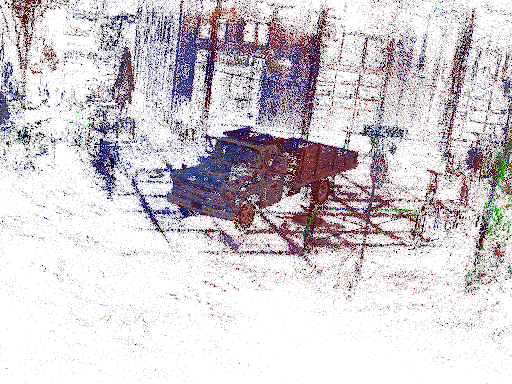}}
    \end{subfigure}
    \hspace{1mm} 
    \begin{subfigure}{0.22\linewidth}
        \centering
        \fbox{\includegraphics[width=\linewidth]{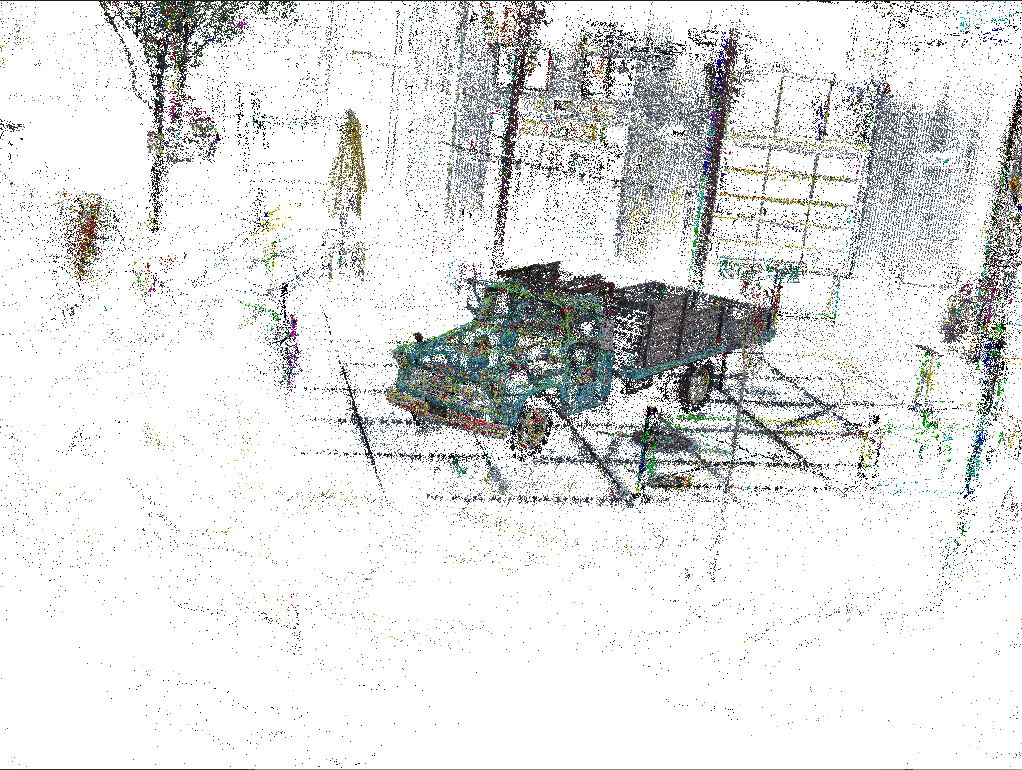}}
    \end{subfigure}


    \begin{subfigure}{0.22\linewidth}
        \centering
        \fbox{\includegraphics[width=\linewidth]{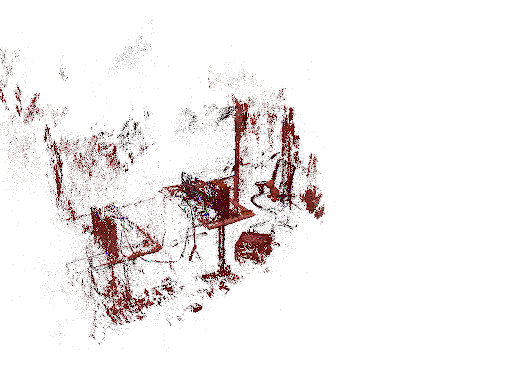}}
    \end{subfigure}
    \hspace{1mm} 
    \begin{subfigure}{0.22\linewidth}
        \centering
        \fbox{\includegraphics[width=\linewidth]{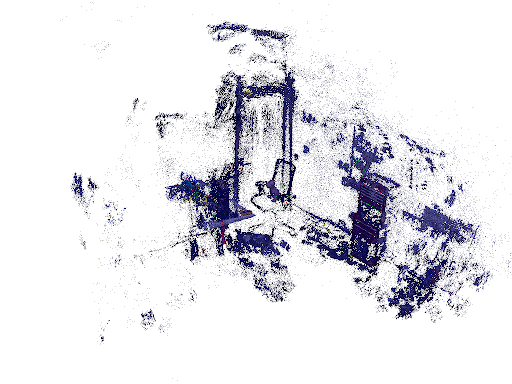}}
    \end{subfigure}
    \hspace{1mm} 
    \begin{subfigure}{0.22\linewidth}
        \centering
        \fbox{\includegraphics[width=\linewidth]{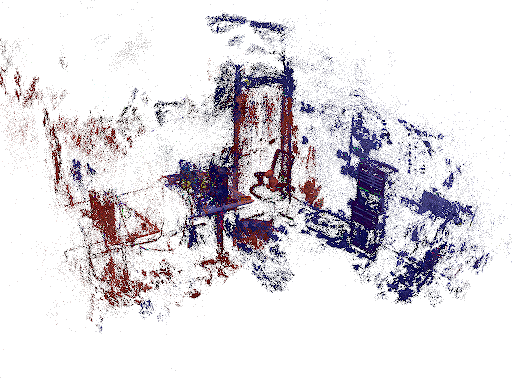}}
    \end{subfigure}
    \hspace{1mm} 
    \begin{subfigure}{0.22\linewidth}
        \centering
        \fbox{\includegraphics[width=\linewidth]{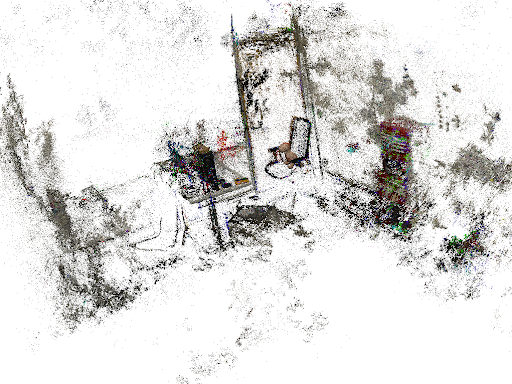}}
    \end{subfigure}


    \begin{subfigure}{0.22\linewidth}
        \centering
        \fbox{\includegraphics[width=\linewidth]{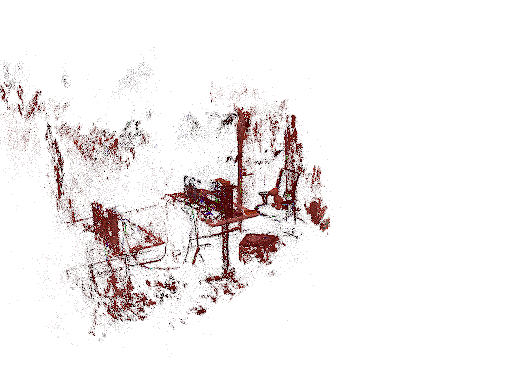}}
        \captionsetup{labelformat=empty}
        \caption{3DGS1}
    \end{subfigure}
    \hspace{1mm} 
    \begin{subfigure}{0.22\linewidth}
        \centering
        \fbox{\includegraphics[width=\linewidth]{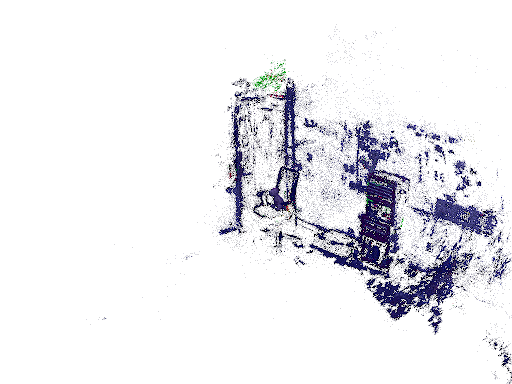}}
        \captionsetup{labelformat=empty}
        \caption{3DGS2}
    \end{subfigure}
    \hspace{1mm} 
    \begin{subfigure}{0.22\linewidth}
        \centering
        \fbox{\includegraphics[width=\linewidth]{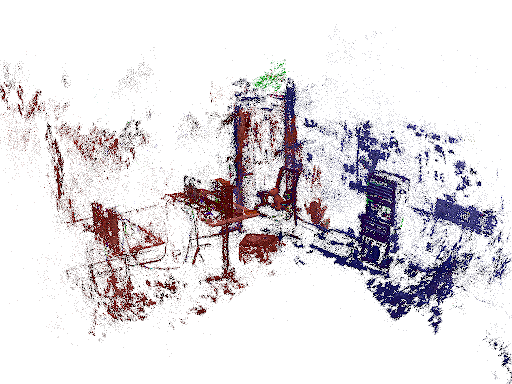}}
        \captionsetup{labelformat=empty}
        \caption{3DGS1 + 3DGS2}
    \end{subfigure}
    \hspace{1mm} 
    \begin{subfigure}{0.22\linewidth}
        \centering
        \fbox{\includegraphics[width=\linewidth]{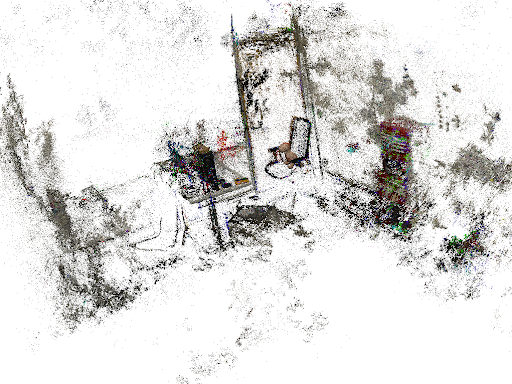}}
        \captionsetup{labelformat=empty}
        \caption{Ground Truth}
    \end{subfigure}
    
    \caption{\ac{3DGS} cloud visualization of PhotoReg registration results. First two columns are input 3DGS; third is our result; last is the Gaussians from the ground truth. We observe PhotoReg tightly aligns the two 3DGS models.}
    \label{pc_viz}
\end{figure}

In this section, we evaluate the performance of the proposed PhotoReg method on merging two or more Gaussian Splatting models with different levels of overlap. The foundation models we used within the framework are {DUSt3R} and {DINOv2}. DUSt3R produces coarse 3D reconstruction from rendered images; DINOv2 performs feature extraction based on robust visual features from images. We seek to empirically answer several major questions:
\begin{enumerate}
\item Can {PhotoReg}, with foundation models, produce accurate registration between two Gaussian Splatting models with partial overlap?
\item How does its performance compare to other classic registration methods?
\item How is the performance of {PhotoReg} on Gaussian Splatting Models with minimal overlap?
\item Can {PhotoReg} handle registration of multiple Gaussian Splatting models?
\item Can PhotoReg be used on image sequences collected on multiple real quadruped robots?
\end{enumerate}

\subsection{Gaussian Splatting Registration with PhotoReg}
\textbf{Dataset:} We use the \emph{Playroom}, \emph{Truck}, and \emph{Train} dataset used in \cite{kerbl3Dgaussians}, in conjunction with those we have gathered independently. Each dataset from \cite{kerbl3Dgaussians} is divided into two subsets. Additionally, we also collect multiple real-world datasets within a room, which we name \emph{Workroom 1} and \emph{Workroom 2}. In Workroom 1, there is very low overlap between the image sequences, while Workroom 2 contains a moderate level of overlap. Each dataset contains independently recorded image sequences. We also have a dataset \emph{Sofa} collected in a different space, using cameras mounted on two different quadrupeds. \textbf{Data Preprocessing:} From each image set, \( 20\% \) of the images are reserved for testing. We denote the rest training sets as $I_1$ and $I_2$ respectively. Gaussian models, $G_1$ and $G_2$ are then trained on $I_1$ and $I_2$. \textbf{Evaluations:} We evaluate the registration quality as photometric loss between rendered images and held-out test images. We use well-established metrics, also used in the NeRF and Splatting Communities \cite{nerf_review_frank, kerbl3Dgaussians}, to evaluate the errors and qualities in the rendered image. These metrics are: SSIM ~\cite{wang2004image} measures structural similarity, PSNR ~\cite{hore2010image} quantifies image fidelity, and LPIPS ~\cite{zhang2018unreasonable} assesses perceptual similarity based on deep neural networks.

\begin{figure}[t]
\centering
\includegraphics[width=\linewidth]{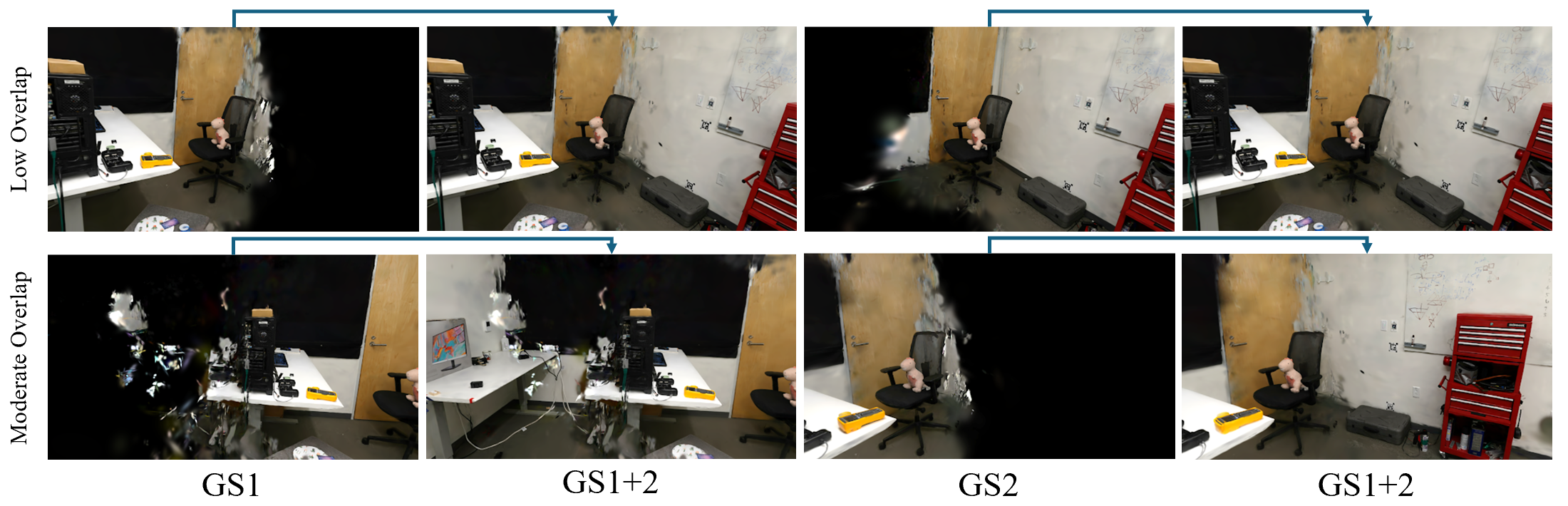}
        \caption{We illustrate results on the workroom 1 and 2 datasets, which have different levels of overlap. From left to right, we show: rendering of G1 at view 1; rendering of fused model at view 1; rendering of G2 at view 2; rendering of fused model at view 2.}
    \label{exp:results2}
\end{figure}

By rendering images from the aligned model, we provide qualitative results on the Tank and Trains dataset in \cref{exp:results1}, and rendered images of our workroom datasets are shown in \cref{exp:results2}. Additionally, visualizations of the 3D Gaussian Clouds of each individual model, and its alignment are shown in \cref{pc_viz}. We observe in each case, PhotoReg provides tight and precise alignments. The alignment quality is highlighted when we zoom (shown by red boxes) into blurry regions of the individual Splatting models --- we observe that the aligned model renders images of much higher visual quality at these regions compared to the initial models.

\subsection{Comparisons with Baseline Methods}

The core at Gaussian Splatting Registration is obtaining the transformation function between target \ac{3DGS} models. Here, we outline competitive baselines that are used to stress-test the performance of PhotoReg: \textbf{Iterative Closest Point (ICP)}
Traditionally, \ac{ICP} is applied for 3D point cloud alignment by iteratively minimizing the distance between corresponding points until convergence. In the case of \ac{3DGS} models, we treat each Gaussian's mean as a point in the cloud. The initial estimation for ICP is provided through DUSt3R, the foundation model. \textbf{COLMAP} \cite{schoenberger2016sfm} is the most widely used Structure-from-Motion method. We can obtain the transformation matrix from camera poses estimated by COLMAP to align each \ac{3DGS} model. 

We compare our method with ICP with scaling enabled, as implemented in Open3D~\cite{du2007icp}, COLMAP, and Ground Truth. Ground Truth is obtained by building a \ac{3DGS} model on all the image sequences used. This represents an upper-bound to the quality of the aligned and merged model. The resultant GS Model from each method is evaluated by metrics previously described. 

The results are tabulated in \cref{tableresult}. We observe that the merged GS model from PhotoReg consistently outperforms the alignment via ICP or COLMAP, and is much closer to the quality of images generated by the ground truth model. Moreover, in the case of low overlap, both ICP and COLMAP fail while PhotoReg accurately aligns.

\subsection{Performance in Low Overlap Scenarios}

Our method, PhotoReg, demonstrates robust performance in scenes with low overlap, where traditional methods often falter. As detailed in \cref{tableresult}, within the Workroom 1 dataset, where there is a low overlap between the separate image sequences, PhotoReg successfully merges the individual \ac{3DGS} models with rendering quality approaching that of the ground truth. In contrast, COLMAP fails to produce any transformation, and ICP results in inaccurate alignments. 
We observe that COLMAP relies heavily on identifying common visual feature points for matching and alignment. When the overlap of the image sequences is low and common visual features cannot be identified, COLMAP struggles to find alignments between the scenes. 
ICP, in contrast, depends on point cloud registration. The overlapping regions are typically on the periphery of the input 3DGS scenes. In low overlap situations, ICP struggles to identify correspondences between the points. {PhotoReg}, our proposed method, circumvents the drawbacks associated with both COLMAP and ICP. Unlike COLMAP, which relies on traditional visual features, PhotoReg leverages foundation models to uncover regions that are visually and semantically correspondent. These data-driven models are invariant to viewing angles, and adept at recovering alignments even when common visual features are sparse or obscured, as is often the case in low overlap or poor-quality image scenarios. This capability demonstrates PhotoReg's resilience and effectiveness, even in conditions of minimal overlap.

\subsection{Multiple Gaussian Splatting Registration}

\begin{figure}[t]
    \centering
    
    \begin{subfigure}{0.48\linewidth}
        \centering
        \includegraphics[width=\linewidth]{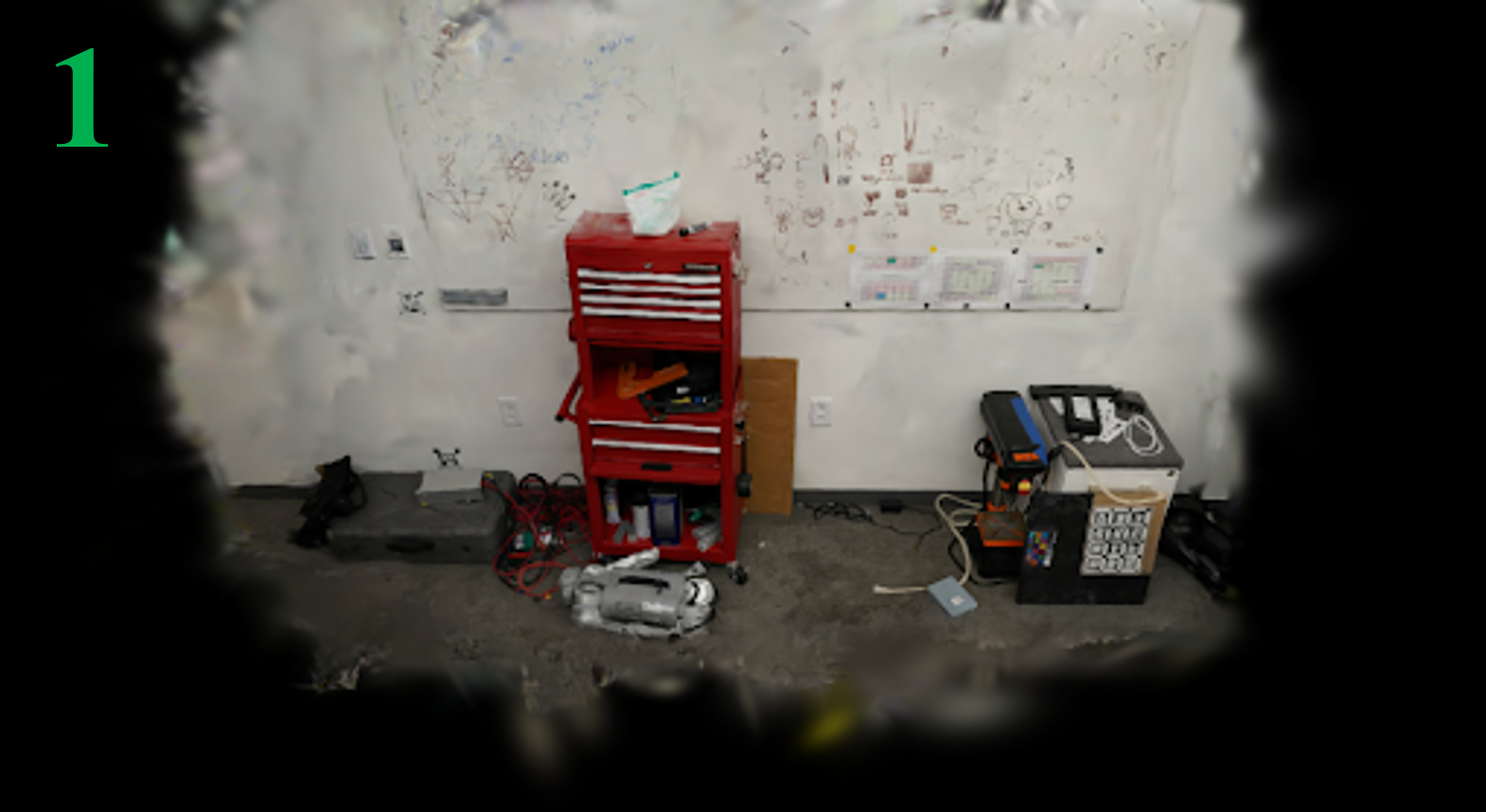}
        \caption{Submodel 1}
    \end{subfigure}
    \begin{subfigure}{0.48\linewidth}
        \centering
        \includegraphics[width=\linewidth]{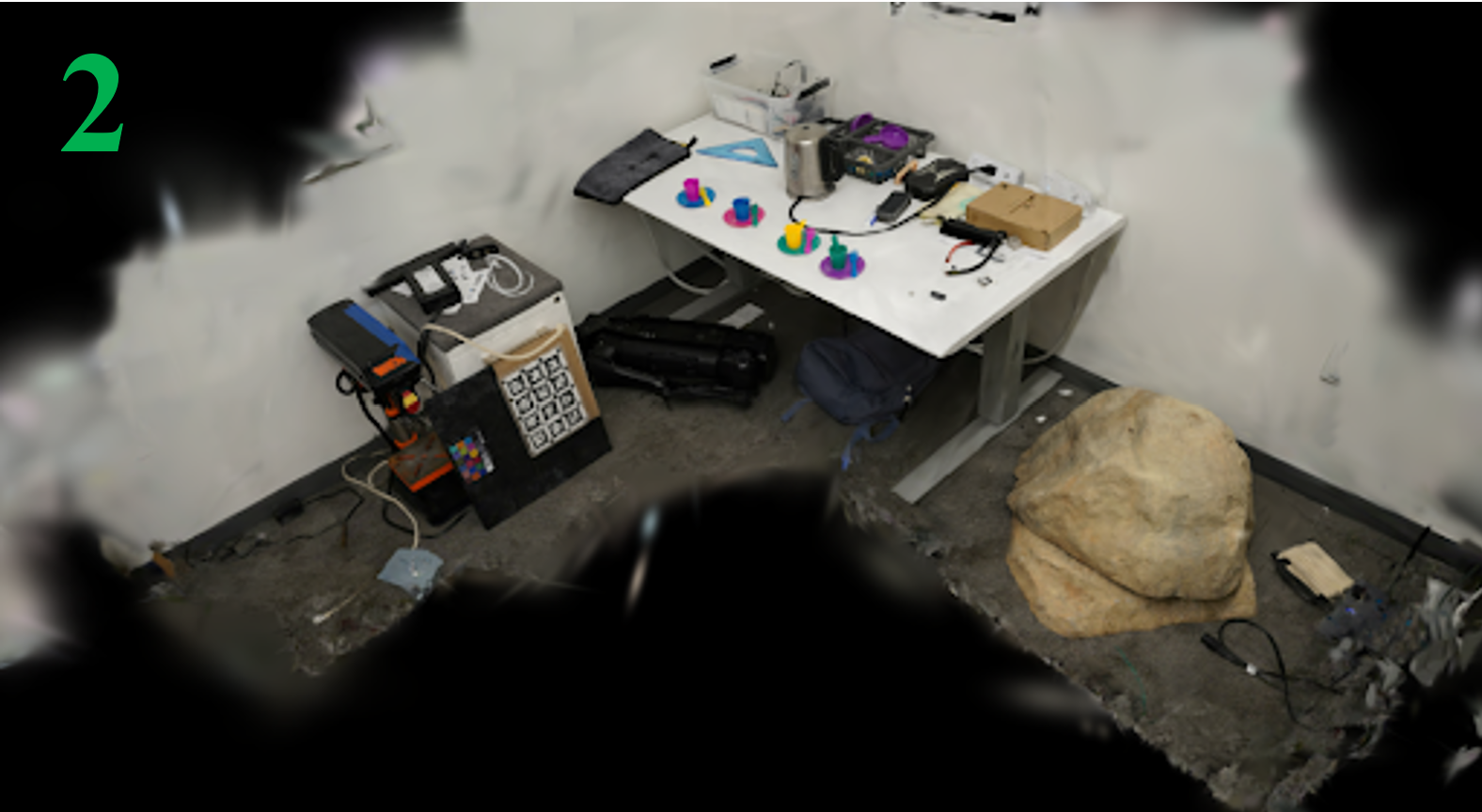}
        \caption{Submodel 2}
    \end{subfigure}

    \begin{subfigure}{0.48\linewidth}
        \centering
        \includegraphics[width=\linewidth]{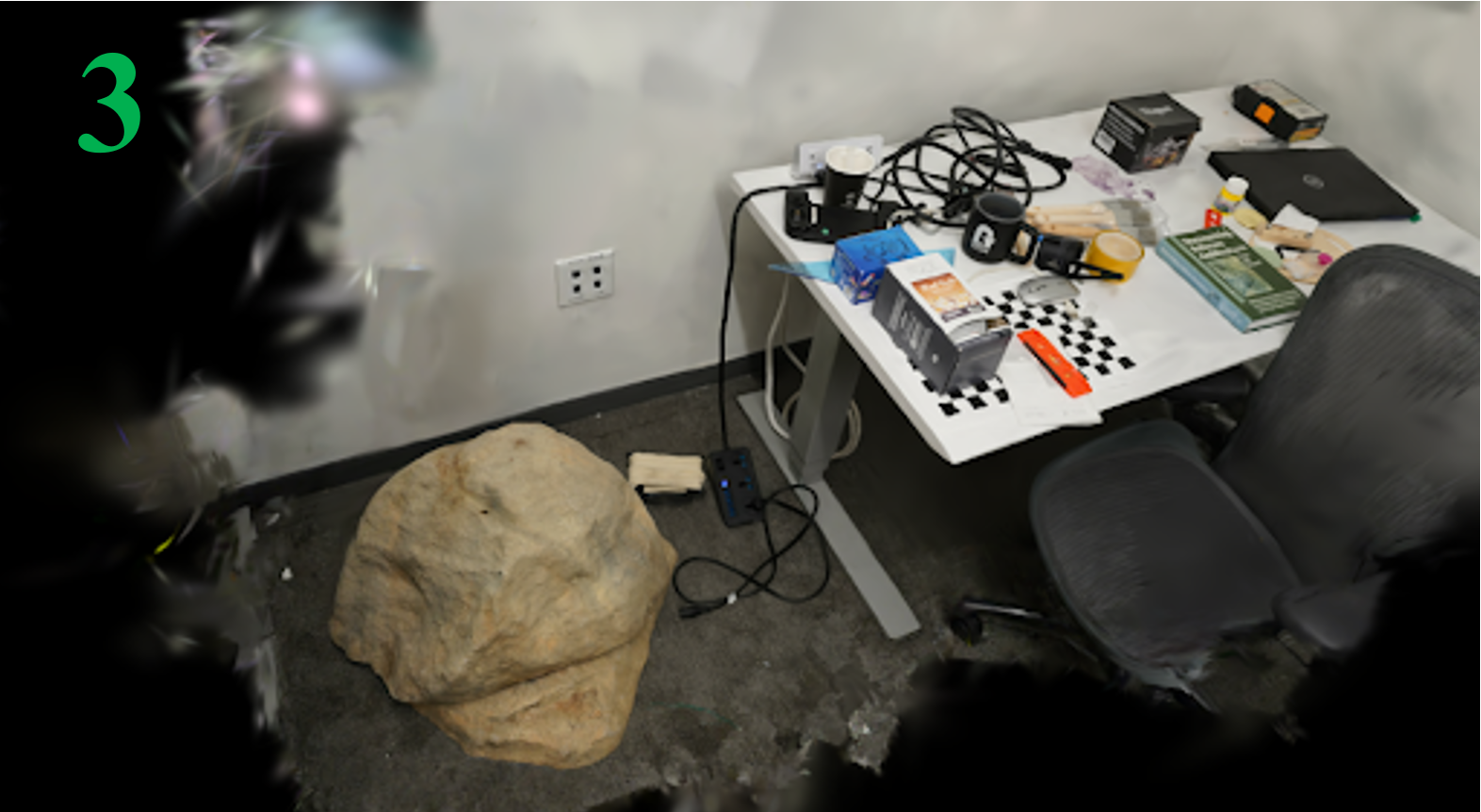}
        \caption{Submodel 3}
    \end{subfigure}
    \begin{subfigure}{0.48\linewidth}
        \centering
        \includegraphics[width=\linewidth]{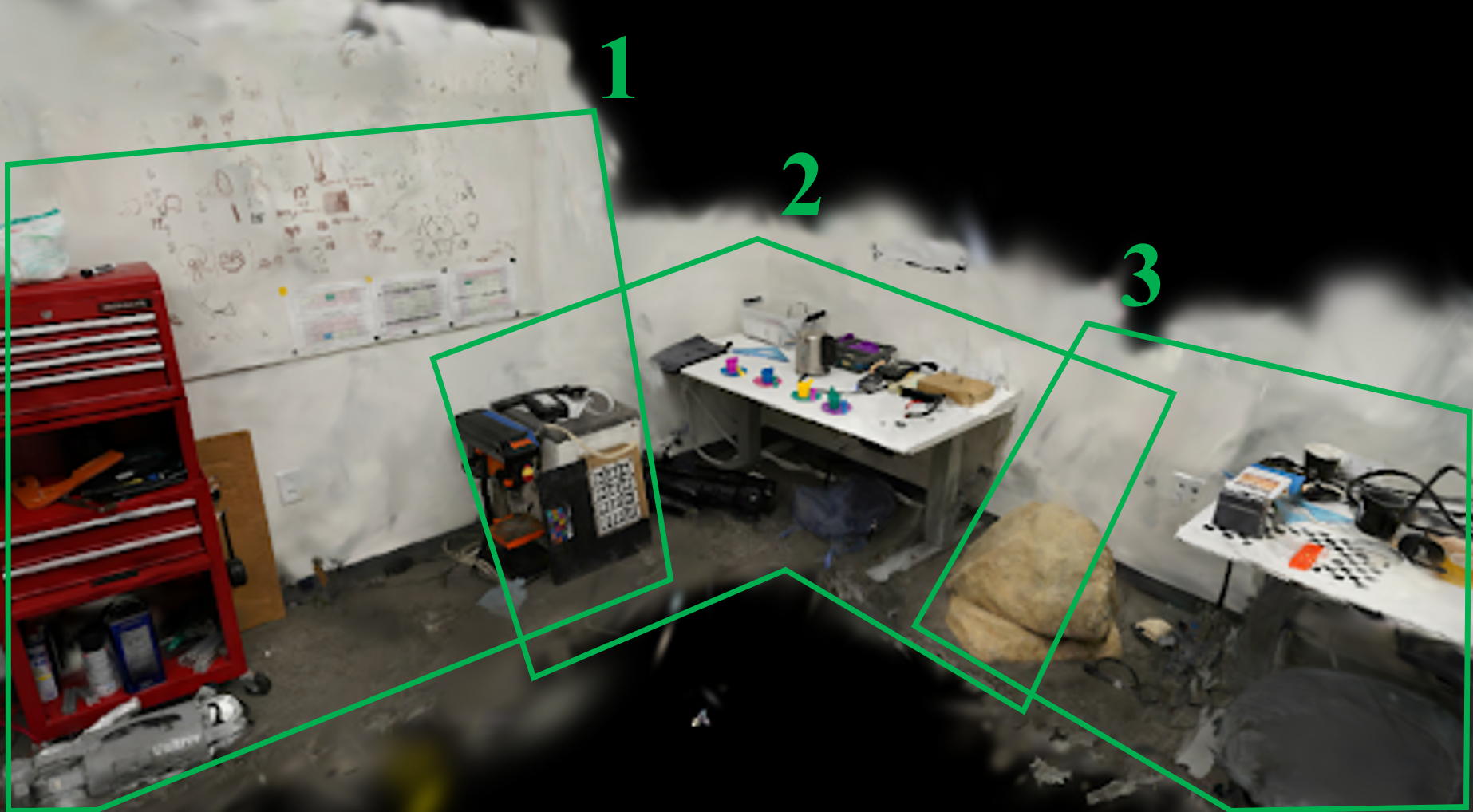}
        \caption{Fused result}
    \end{subfigure}

    \caption{Visualization of multiple 3DGS registration results. Image (a), (b), (c) are 3DGS Models we want to align together. Image (d) shows the resultant 3DGS model, with green boxes in the fused result outlining the location of the shown individual models.}
    \label{multi}
\end{figure}

In many scenarios, such as robotic mapping of unknown environments, teams of more than two robots may be deployed. Each robot contributes to the mapping process by generating its own \ac{3DGS} model, necessitating the integration of multiple 3DGS models. We extend our experiment to merging more than two GS models. As shown in \cref{multi}, we collect three \ac{3DGS} models depicting three corners in a room. the resulting \ac{3DGS} scene correctly aligned all three models together into a complete and coherent scene. This demonstrates our method's robustness in handling complex, multi-robot mapping tasks.

\begin{figure}[t]%
\centering
\includegraphics[width=\linewidth]{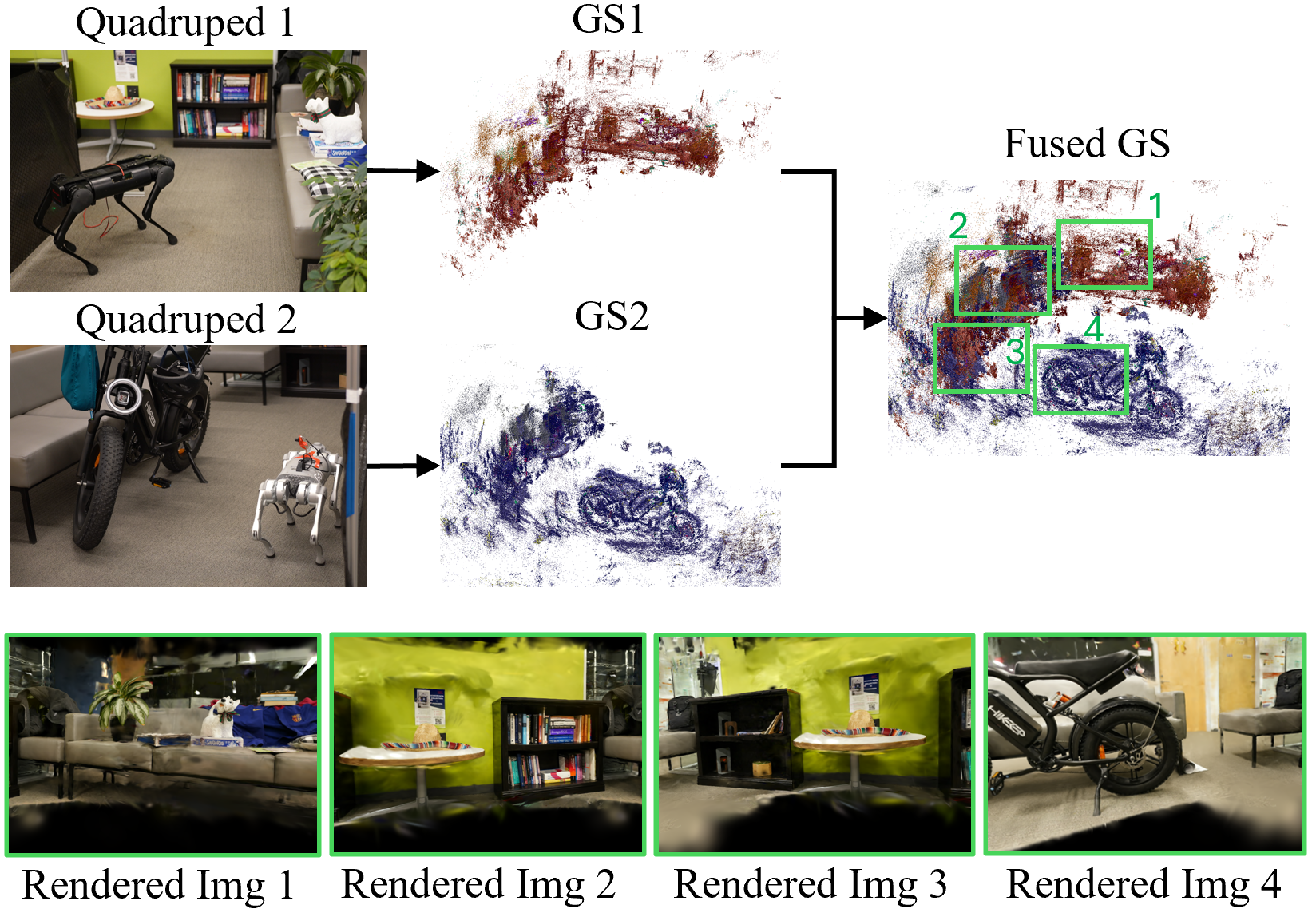}
    \caption{Fused result on robot-collected data. Two quadrupeds are deployed to collect real-world data. The resulting \ac{3DGS} clouds and fused \ac{3DGS} clouds are shown in the figure. Four images are also rendered with the Fused GS. The spatial locations of the rendered images are outlined in green. We observe that the rendering is of high quality at the overlapping regions.}
    \label{roboexp}
\end{figure}

\subsection{PhotoReg on Robot-collected Data}
We applied our PhotoReg framework to data collected by two quadruped robots equipped with USB cameras. Each robot was assigned to scan a separate side of the scene. To separate the two regions, a black canvas was placed in the middle, blocking the view between the left and right sides, while still allowing some overlap in shared areas, such as bookshelves. Using the images captured by each quadruped, we generated individual \ac{3DGS} models. These models were then processed through our PhotoReg framework to produce a fused GS. The resulting fused model, with rendered images and an illustration of the Gaussian clouds before and after the alignment are illustrated in \cref{roboexp}. The merged \ac{3DGS} generates high-quality images at regions of overlap.




\section{CONCLUSIONS AND FUTURE WORK}
\label{sec:conclusions}
We present a novel framework, PhotoReg, to register photorealistic Gaussian Splatting models. To tackle this, we advocate for applying pre-trained {foundation models} to produce correspondences and an initial alignment. We develop methodologies to resolve scale disparities between the individual models. Then, PhotoReg fine-tunes the results by optimizing for a precise alignment using photometric losses. We demonstrate by doing the accuracy and robustness of our results on a variety of benchmark datasets and also collected real-world data. In particular, we empirically demonstrate that PhotoReg outshines competitive baselines when aligning individual models with little overlap. Future research avenues include (1) considering the fusion dynamic Gaussian Splatting models which vary over time. (2) Considering multiple robot problems to actively build a photorealistic environment, with certain photometric requirements in different regions of the environment.

\bibliographystyle{ieeetr}
\bibliography{ref.bib}

\begin{thebibliography}{10}

\bibitem{agha2021nebula}
A.~Agha, K.~Otsu, B.~Morrell, D.~D. Fan, R.~Thakker, A.~Santamaria-Navarro, S.-K. Kim, A.~Bouman, X.~Lei, J.~Edlund, {\em et~al.}, ``Nebula: Quest for robotic autonomy in challenging environments; team costar at the darpa subterranean challenge,'' {\em arXiv preprint arXiv:2103.11470}, 2021.

\bibitem{keetha2023anyloc}
N.~Keetha, A.~Mishra, J.~Karhade, K.~M. Jatavallabhula, S.~Scherer, M.~Krishna, and S.~Garg, ``Anyloc: Towards universal visual place recognition,'' {\em IEEE Robotics and Automation Letters}, 2023.

\bibitem{zhang2023nerf}
T.~Zhang and M.~Johnson-Roberson, ``Beyond nerf underwater: Learning neural reflectance fields for true color correction of marine imagery,'' {\em IEEE Robotics and Automation Letters}, 2023.

\bibitem{zhang2024darkgs}
T.~Zhang, K.~Huang, W.~Zhi, and M.~Johnson-Roberson, ``Darkgs: Learning neural illumination and 3d gaussians relighting for robotic exploration in the dark,'' 2024.

\bibitem{zhang2021self}
Y.~Zhang, J.~D. Lee, and Q.~Liu, ``Self-supervised learning and pre-training in generalized and natural language task,'' {\em arXiv preprint arXiv:2103.02690}, 2021.

\bibitem{zhou2016fast}
Q.-Y. Zhou, J.~Park, and V.~Koltun, ``Fast global registration,'' in {\em European Conference on Computer Vision (ECCV)}, pp.~766--782, Springer, 2016.

\bibitem{yuan2020self}
Y.~Yuan, J.~Hou, A.~Nüchter, and S.~Schwertfeger, ``Self-supervised point set local descriptors for point cloud registration,'' {\em arXiv preprint arXiv:2003.05199}, 2020.

\bibitem{du2007icp}
S.~Du, N.~Zheng, S.~Ying, Q.~You, and Y.~Wu, ``An extension of the icp algorithm considering scale factor,'' in {\em 2007 IEEE International Conference on Image Processing}, (San Antonio, TX, USA), pp.~V -- 193--V -- 196, IEEE, 2007.

\bibitem{PDMP}
T.~Lai, W.~Zhi, T.~Hermans, and F.~Ramos, ``Parallelised diffeomorphic sampling-based motion planning,'' in {\em Conference on Robot Learning (CoRL)}, 2021.

\bibitem{GeoFab_gloabL_opt}
W.~Zhi, I.~Akinola, K.~van Wyk, N.~Ratliff, and F.~Ramos, ``Global and reactive motion generation with geometric fabric command sequences,'' in {\em IEEE International Conference on Robotics and Automation, ICRA}, 2023.

\bibitem{Diff_templates}
W.~Zhi, T.~Lai, L.~Ott, and F.~Ramos, ``Diffeomorphic transforms for generalised imitation learning,'' in {\em Learning for Dynamics and Control Conference, {L4DC}}, 2022.

\bibitem{OccupancyGridMaps}
A.~Elfes, ``Sonar-based real-world mapping and navigation,'' {\em IEEE Journal on Robotics and Automation}, 1987.

\bibitem{HM}
W.~{Zhi}, L.~{Ott}, R.~{Senanayake}, and F.~{Ramos}, ``Continuous occupancy map fusion with fast bayesian hilbert maps,'' in {\em International Conference on Robotics and Automation (ICRA)}, 2019.

\bibitem{DirectionalGridMaps}
R.~Senanayake and F.~Ramos, ``Directional grid maps: modeling multimodal angular uncertainty in dynamic environments,'' in {\em IEEE/RSJ International Conference on Intelligent Robots and Systems}, 2018.

\bibitem{sptemp}
W.~{Zhi}, R.~{Senanayake}, L.~{Ott}, and F.~{Ramos}, ``Spatiotemporal learning of directional uncertainty in urban environments with kernel recurrent mixture density networks,'' {\em IEEE Robotics and Automation Letters}, 2019.

\bibitem{OTNet}
W.~Zhi, T.~Lai, L.~Ott, and F.~Ramos, ``Trajectory generation in new environments from past experiences,'' in {\em IEEE/RSJ International Conference on Intelligent Robots and Systems (IROS)}, 2021.

\bibitem{KTM}
W.~Zhi, L.~Ott, and F.~Ramos, ``Kernel trajectory maps for multi-modal probabilistic motion prediction,'' in {\em Conference on Robot Learning (CoRL)}, 2019.

\bibitem{mildenhall2020nerf}
B.~Mildenhall, P.~P. Srinivasan, M.~Tancik, J.~T. Barron, R.~Ramamoorthi, and R.~Ng, ``Nerf: Representing scenes as neural radiance fields for view synthesis,'' in {\em ECCV}, 2020.

\bibitem{mueller2022instant}
T.~M\"uller, A.~Evans, C.~Schied, and A.~Keller, ``Instant neural graphics primitives with a multiresolution hash encoding,'' {\em ACM Trans. Graph.}, 2022.

\bibitem{kerbl3Dgaussians}
B.~Kerbl, G.~Kopanas, T.~Leimk{\"u}hler, and G.~Drettakis, ``3d gaussian splatting for real-time radiance field rendering,'' {\em ACM Transactions on Graphics}, vol.~42, no.~4, 2023.

\bibitem{zhang2024recgs}
T.~Zhang, W.~Zhi, K.~Huang, J.~Mangelson, C.~Barbalata, and M.~Johnson-Roberson, ``Recgs: Removing water caustic with recurrent gaussian splatting,'' {\em arXiv preprint arXiv:2407.10318}, 2024.

\bibitem{chen1992objectICP}
Y.~Chen and G.~Medioni, ``Object modelling by registration of multiple range images,'' {\em Image and vision computing}, vol.~10, no.~3, pp.~145--155, 1992.

\bibitem{park2017colored}
J.~Park, Q.-Y. Zhou, and V.~Koltun, ``Colored point cloud registration revisited,'' in {\em Proceedings of the IEEE International Conference on Computer Vision (ICCV)}, pp.~143--152, IEEE, 2017.

\bibitem{besl1992method}
P.~J. Besl and N.~D. McKay, ``A method for registration of 3-d shapes,'' {\em IEEE Transactions on Pattern Analysis and Machine Intelligence}, vol.~14, no.~2, pp.~239--256, 1992.

\bibitem{jian2011robust}
B.~Jian and B.~C. Vemuri, ``Robust point set registration using gaussian mixture models,'' {\em IEEE Transactions on Pattern Analysis and Machine Intelligence}, vol.~33, no.~8, pp.~1633--1645, 2011.

\bibitem{goli2023nerf2nerf}
L.~Goli, D.~Rebain, S.~Sabour, A.~Garg, and A.~Tagliasacchi, ``nerf2nerf: Pairwise registration of neural radiance fields,'' in {\em International Conference on Robotics and Automation (ICRA)}, IEEE, 2023.

\bibitem{DReg2023}
Y.~Chen and G.~H. Lee, ``Dreg-nerf: Deep registration for neural radiance fields,'' {\em 2023 {IEEE/CVF} International Conference on Computer Vision}, 2023.

\bibitem{zhu2024loopsplat}
L.~Zhu, Y.~Li, E.~Sandström, S.~Huang, K.~Schindler, and I.~Armeni, ``Loopsplat: Loop closure by registering 3d gaussian splats,'' {\em arXiv preprint arXiv:2408.10154}, 2024.

\bibitem{Bommasani2021FoundationModels}
R.~Bommasani and et~al., ``On the opportunities and risks of foundation models,'' {\em CoRR}, 2021.

\bibitem{chen2020simple}
T.~Chen, S.~Kornblith, M.~Norouzi, and G.~Hinton, ``A simple framework for contrastive learning of visual representations,'' in {\em International Conference on Machine Learning}, pp.~1597--1607, PMLR, 2020.

\bibitem{donahue2019large}
J.~Donahue and K.~Simonyan, ``Large scale adversarial representation learning,'' in {\em Advances in Neural Information Processing Systems}, vol.~32, 2019.

\bibitem{grill2020bootstrap}
J.-B. Grill, F.~Strub, F.~Altch{\'e}, C.~Tallec, P.~H. Richemond, E.~Buchatskaya, C.~Doersch, B.~A. Pires, Z.~D. Guo, M.~G. Azar, B.~Piot, K.~Kavukcuoglu, R.~Munos, and M.~Valko, ``Bootstrap your own latent: A new approach to self-supervised learning,'' in {\em Advances in Neural Information Processing Systems (NeurIPS)}, pp.~21271--21284, 2020.

\bibitem{oquab2023dinov2}
M.~Oquab, T.~Darcet, T.~Moutakanni, H.~V. Vo, M.~Szafraniec, V.~Khalidov, P.~Fernandez, D.~Haziza, F.~Massa, A.~El-Nouby, R.~Howes, P.-Y. Huang, H.~Xu, V.~Sharma, S.-W. Li, W.~Galuba, M.~Rabbat, M.~Assran, N.~Ballas, G.~Synnaeve, I.~Misra, H.~Jegou, J.~Mairal, P.~Labatut, A.~Joulin, and P.~Bojanowski, ``Dinov2: Learning robust visual features without supervision,'' 2023.

\bibitem{DUSt3R_cvpr24}
S.~Wang, V.~Leroy, Y.~Cabon, B.~Chidlovskii, and J.~Revaud, ``Dust3r: Geometric 3d vision made easy,'' in {\em CVPR}, 2024.

\bibitem{JCR}
W.~Zhi, H.~Tang, T.~Zhang, and M.~Johnson-Roberson, ``Unifying representation and calibration with 3d foundation models,'' {\em IEEE Robotics and Automation Letters}, 2024.

\bibitem{zhi20243d}
W.~Zhi, H.~Tang, T.~Zhang, and M.~Johnson-Roberson, ``3d foundation models enable simultaneous geometry and pose estimation of grasped objects,'' {\em arXiv preprint arXiv:2407.10331}, 2024.

\bibitem{dosovitskiy2020vit}
A.~Dosovitskiy, L.~Beyer, A.~Kolesnikov, D.~Weissenborn, X.~Zhai, T.~Unterthiner, M.~Dehghani, M.~Minderer, G.~Heigold, S.~Gelly, J.~Uszkoreit, and N.~Houlsby, ``An image is worth 16x16 words: Transformers for image recognition at scale,'' {\em ICLR}, 2021.

\bibitem{gaussiansplatl2024}
Z.~Yuan, ``{Gaussian Splatting Lightning}.'' \url{https://github.com/yzslab/gaussian-splatting-lightning}, 2024.

\bibitem{zhang2024radegs}
B.~Zhang, C.~Fang, R.~Shrestha, Y.~Liang, X.~Long, and P.~Tan, ``Rade-gs: Rasterizing depth in gaussian splatting,'' {\em arXiv preprint arXiv:2406.01467}, 2024.

\bibitem{nerf_review_frank}
F.~Dellaert and Y.~Lin, ``Neural volume rendering: Nerf and beyond,'' {\em CoRR}, 2021.

\bibitem{wang2004image}
Z.~Wang, A.~C. Bovik, H.~R. Sheikh, and E.~P. Simoncelli, ``Image quality assessment: from error visibility to structural similarity,'' {\em IEEE Transactions on Image Processing}, vol.~13, no.~4, pp.~600--612, 2004.

\bibitem{hore2010image}
A.~Horé and D.~Ziou, ``Image quality metrics: Psnr vs. ssim,'' in {\em 2010 International Conference on Pattern Recognition (ICPR)}, pp.~2366--2369, IEEE, 2010.

\bibitem{zhang2018unreasonable}
R.~Zhang, P.~Isola, A.~A. Efros, E.~Shechtman, and O.~Wang, ``The unreasonable effectiveness of deep features as a perceptual metric,'' {\em arXiv preprint arXiv:1801.03924}, 2018.

\bibitem{schoenberger2016sfm}
J.~L. Sch\"{o}nberger and J.-M. Frahm, ``Structure-from-motion revisited,'' in {\em Conference on Computer Vision and Pattern Recognition (CVPR)}, 2016.

\end{thebibliography}
\end{document}